# Multiple receptive fields and small-object-focusing weakly-supervised segmentation network for fast object detection

Siyang Sun[1,2,3], Yingjie Yin[1,2], Xingang Wang[1,2], De Xu[1,2], Yuan Zhao[3], Haifeng Shen[3]

[1]Research Center of Precision Sensing and Control, Institute of Automation, Chinese Academy of Sciences, Beijing 100190, China
[2]School of Artificial Intelligence, University of Chinese Academy of Science, Beijing 100190, China
[3]AI Labs, Didi Chuxing, Beijing 100193, China
{sunsiyang2015, yingjie.yin, xingang.wang, de.xu}@ia.ac.cn
{zhaoyuanjason, shenhaifeng}@didiglobal.com

*Abstract*— Object detection plays an important role in various visual applications. However, the precision and speed of detector are usually contradictory. One main reason for fast detectors' precision reduction is that small objects are hard to be detected. To address this problem, we propose a multiple receptive field and small-object-focusing weakly-supervised segmentation network (MRFSWSnet) to achieve fast object detection. In MRFSWSnet, multiple receptive fields block (MRF) is used to pay attention to the object and its adjacent background's different spatial location with different weights to enhance the feature's discriminability. In addition, in order to improve the accuracy of small object detection, a small-object-focusing weakly-supervised segmentation module which only focuses on small object instead of all objects is integrated into the detection network for auxiliary training to improve the precision of small object detection. Extensive experiments show the effectiveness of our method on both PASCAL VOC and MS COCO detection datasets. In particular, with a lower resolution version of 300×300, MRFSWSnet achieves 80.9% mAP on VOC2007 *test* with an inference speed of 15 milliseconds per frame, which is the state-of-the-art detector among real-time detectors.

## I. INTRODUCTION

OBJECT detection plays an important role in many visual applications, such as visual navigation [1]-[2], video surveillance [3]-[4], intelligent transport [5]-[6] and so on. Minaeian [1] proposed a customized detection algorithm for UAV's navigation, and Yuan [2] presented a novel context-aware multichannel feature pyramid for vehicle's navigation. Yang [5] also proposed a fast and accurate vanishing point detection method for various types of roads used for autopilot. Comparing with traditional object detection methods based on hand-craft features [7]-[11], recent detectors based on deep convolution neural network (CNN) [12]-[24] show powerful performance because of robust and discriminate features. The CNN-based methods for object detection can be divided into two classes, which are region-based two-stage detector [12]-[17] and region-free one-stage detector [18]-[24]. Two-stage detectors [12]-[17] achieve higher precision, however, their complex computation and lower speed limited the practical application. To accelerate the speed of object detection, several one-stage detectors [18]-[24] were proposed. Their running speed keep real time performance, but the accuracy has a clear drop which is about 10% to 40% relative to state-of-the-art two-stage detector [12]-[17]. Previous one-stage detector, such as SSD [19], designed a serial of reference anchor boxes with different scales and aspect ratios and directly regressed these anchors on features from different levels. Lower features are mainly used to detect smaller object and higher features are used to detect larger object. However, the lower feature with less semantic information will result in difficult detection for small objects. In order to solve the problem of small object detection, several strategies [25]-[27], such as the fusion of multiple scales' features, the new regulation of anchors, the new module of object detector, are introduced into one-stage detector, however, the running speed decreased because of new computation burden. To solve these problems, we proposed multiple receptive fields and small-object-focusing weakly-supervised segmentation network (MRFSWSnet) for fast object detection.

Many advances [28]-[34] have approved the fact that detectors can achieved better result because of receptive field modules. Inception module of GoogleNet [28] uses multiple convolution layers with different kernels which are sampled at the same center to construct its receptive field module. The receptive field block (RFB) [34] was also proposed by imitating the human population receptive field. Different from the Inception module [28], multiple convolution layers with different dilate rates were used to construct RFB. Different from above two types of receptive fields modules used in object detection, multiple receptive fields block (MRF) is designed in the proposed detector, which is composed of multiple convolution layers with different kernels and different dilated rates. The proposed MRF are used to pay attention to the object and its adjacent background's different spatial locations with different weights to enhance the features' discriminability, and it can cover more positions from the center of an object to its surrounding. MRF can achieve better feature discriminability with lower computation burden.

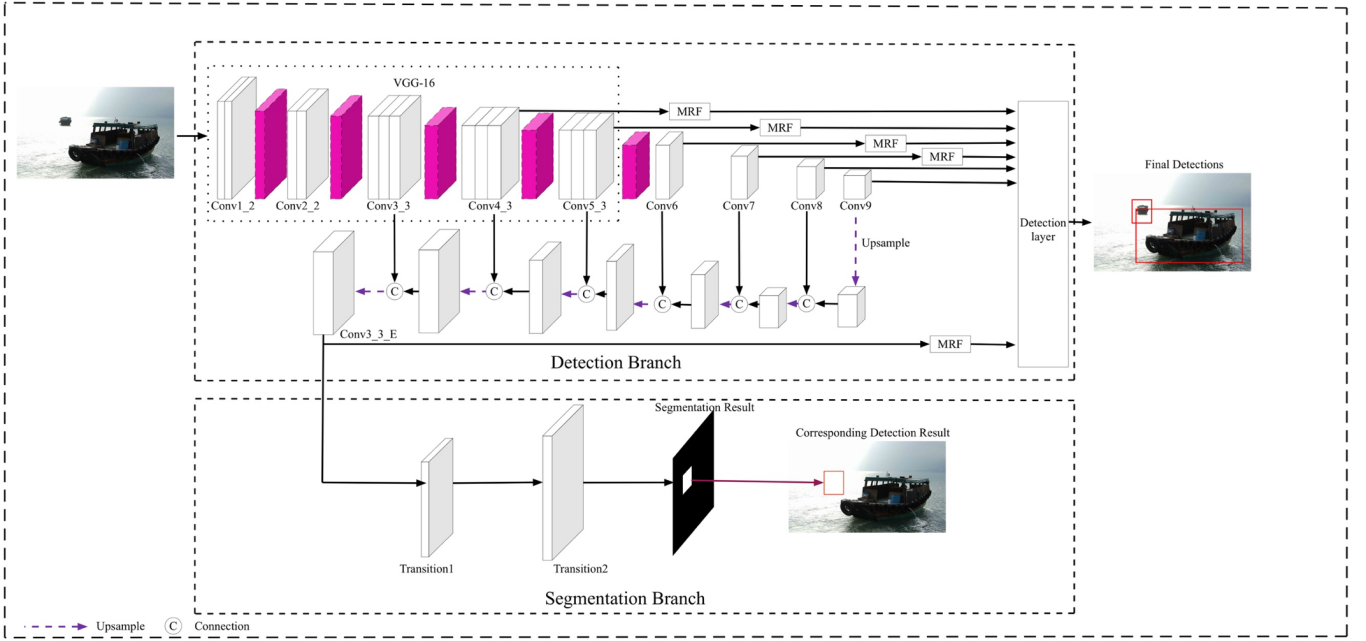

Fig. 1. The structure of the proposed MRFSWSnet

Object detection usually focuses on whole object but lack of paying attention to local cues, while semantic segmentation pays close attention to each position within the object. Many advances [35]-[36] show that combining object detection with weakly-supervised semantic segmentation can improve the performance of object detection. In order to further improve the precision of small object detection, a small-object-focusing weakly-supervised segmentation module (SWS) is integrated into the detection network for auxiliary training.

Our main contributions can be summarized as follows:

(1) Firstly, a multiple receptive field and small-object-focusing weakly-supervised segmentation network (MRFSWSnet) is proposed to achieve fast object detection. It achieves state-of-the-art accuracy with real-time running speed on PASCAL VOC and MS COCO dataset.

(2) Secondly, multiple receptive fields block (MRF) is designed as new convolutional predictors for SSD [19] to improve detection accuracy. MRF are used to pay attention to the object and its adjacent background's different spatial location with different weights to enhance the features' discriminability, and it can cover more positions from the center of an object to its surrounding.

(3) Thirdly, a small-object-focusing weakly-supervised segmentation module is integrated into MRFSWSnet as an auxiliary training task which only focuses on small object instead of all objects to improve the precision of small object detection. MRFSWSnet shows significant performance boosting for small object detection, while it still keeps real-time speed.

## II. RELATED WORK

The CNN-based modern object detector can be divided into two classes, which are region-based two-stage detectors and region-free one-stage detectors. Two-stage detector due to higher accuracy is received attention, the process of which generates a sparse of proposals firstly and then classifies these proposals by classifier. The representative method of two-stage detector is Faster RCNN [14], which generates a serial of candidate proposals by region proposal network (RPN) and then regresses and classifies these proposals through Fast RCNN [13]. Its descendants such as R-FCN [15], FPN [16] and Mask R-CNN [37] are proposed to further improve the detection accuracy. However, their slower running speed and complex computation limited its practical application. One-stage detector is proposed to solve the problem of slower speed and complex computation of two-stage detector. It discards the phase of generating proposals and detects objects in a dense manner e.g. YOLO [18] and SSD [19]. YOLO and SSD adopted lightweight network as backbone to obtain faster running speed at expense of detection accuracy. However, many advances [23] [37] have approved the fact that detectors can achieved better result because of the robust and discriminate features of deeper network [38] with complex computation.

In order to improve performance of detector with less computation, some works focus on enhancing the discrimination of lightweight network's features by different receptive fields. Inception family [28]-[30] were born, which integrated multiple convolution layers with different kernels to get different scales' information with different receptive field. Different from Inception family [28]-[30], ASPP [31] for semantic segmentation adopted multiple dilated convolution layers to generate a serial of uniform resolution features without additional parameters. Deformable CNN [32] proposed a novel dilated convolution operator by learning the offset of individual object on the basic of ASPP. RFB [34] designed a receptive field block with different dilation convolution layers only focusing on special position without focusing all surrounding specified position of one pixel. We propose multiple receptive fields block (MRF) in our detector, which is composed of multiple convolution layers with different kernels and different dilated rates. The proposed MRF pays attention to the different spatial location with different weights according to different distances from objects' center position to its surrounding, and it can cover more surrounding positions of the object.

With these improved methods, the precision for object detection can be promoted. However, the accuracy for small object detection was still lower. In order to improve the accuracy of small object detection, Zhu [26] proposed a novel anchor strategy to support anchor-based face detector, which improve the performance of tiny face detection. Levi [27] constructed the relational module through the spatial and semantic relations for object detection to improve the performance of small object detection. In our detector, a small-object-focusing weakly-supervised segmentation module (SWS) is integrated into the detection network which only focusing on small object for auxiliary training to improve precision of small object detection. Object detection usually focuses on the whole object without paying attention to local cues, while semantic segmentation pays close attention to each position within the object. Many advances [35]-[36] show that combining object detection with weakly semantic segmentation can improve performance of object detection. S. Gidaris et al. [35] proposed a semantic segmentation-aware CNN model for object detection by enhancing the detection feature with semantic segmentation tasks at highest level. The semantic features with weakly-supervised were used to activate the feature of detection. Zhang et al. [36] used the semantic segmentation's feature for activating the feature of detection at lowest layer, which could also improve the detection accuracy. The two methods used the feature of weakly-supervised semantic segmentation to activate the feature of detection, which enhanced the robustness and discrimination of feature for detection. All the objects with weakly-supervised ground truth were used to train the semantic features. Different from the above methods, a small-object-focusing weakly-supervised segmentation module (SWS) is integrated into the detection network which only focusing on small object for auxiliary training. SWS shows significant performance boosting for small object detection, while our detector still keeps real-time speed.

III. MULTIPLE RECEPTIVE FIELDS AND SMALL-OBJECT-FOCUSING WEAKLY-SUPERVISED SEGMENTATION NETWORK

In this section, we firstly introduce the whole detection architecture of MRFSWSnet in Section III.A. Then the multiple receptive fields block (MRF) for object detection is presented in Section III.B. Afterwards, the small-object-focusing weakly-supervised segmentation module (SWS) is described in Section III.C. Finally, the training of MRFSWSnet in the form of multiple tasks is presented in Section III.D.

*A. The network architecture of MRFSWSnet*

The proposed multiple receptive fields and small-object-focusing weakly-supervised segmentation network (MRFSWSnet) is composed of two branches, which are the detection branch and the segmentation branch. The whole network architecture is shown in Fig.1. The detection branch reuses the structure of SSD due to the effectiveness of detection's accuracy and running speed. The base network in MRFSWSnet is same with original SSD, which is VGG-16 [39] trained in ImageNet dataset [40].

In our detection branch of MRFSWSnet, we keep the cascade structure unchanged while add one feature conv3_3_E generated by feature pyramid network (FPN) [16] in our MRFSWSnet. FPN [16] has been proven to ameliorate the performance of features and improve the detection accuracy. With a coarser resolution's feature map conv9 as shown in Fig. 1, we upsample the previous feature by a factor of 2 (using nearest neighbor upsampling), and then the upsampled feature is merged with the corresponding bottom-up feature map. The process does not end until the last finest-resolution feature map conv3_3_E is obtained, which is used as the first detection feature. Different from the original SSD, seven features (conv9, conv8, conv7, conv6, conv5_3, conv4_3, conv3_3_E) as shown in Fig. 1 are used as predicted features and each one except for conv9 and conv8 follows a multiple receptive fields (MRF) block as a new convolutional predictor. The resolution of last two detection feature maps are 1×1 and 3×3, which is unable to be replaced with MRF because these feature maps are too smaller to apply the larger kernels with 5×5 size. The detail structure of MRF will be shown in Section III.B.

In our segmentation branch of MRFSWSnet, an auxiliary small-object-focusing weakly-supervised semantic segmentation module (SWS) follows conv3_3_E. SWS focuses only small object instead of all objects and it is integrated into the detection network for auxiliary training to improve the precision of small object detection. The detection branch and segmentation branch are combined in the form of multiple tasks.

*B. Multiple Receptive Fields Block(MRF)*

As shown in Fig. 2, the proposed multiple receptive fields block (MRF) is composed of multiple convolution layers with different receptive fields for focusing on more positions from the center of an object to its surrounding, and these convolution layers can be divided into two parts according to the forms of different receptive fields. One part of MRF is similar with Inception structure [28], whose components are several convolution layers with different kernel sizes, including 1×1 convolution layer, 3×3 convolution layer and 5×5 convolution layer. The receptive fields of these convolution layers can cover more positions of one object from its center to its surrounding. The other part of MRF includes several dilated convolution layers with different dilated ratios, such as 3×3 convolution layer with dilated ratios 1, 2 and 3, respectively. These convolution layers pay attention to different spatial locations with different weights according to different distances from objects' center position to its surrounding.

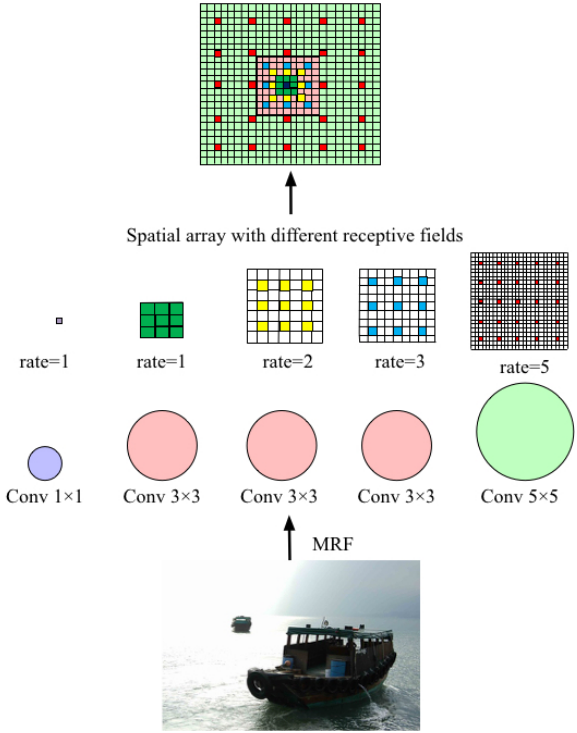

Fig. 2. The multiple receptive field block

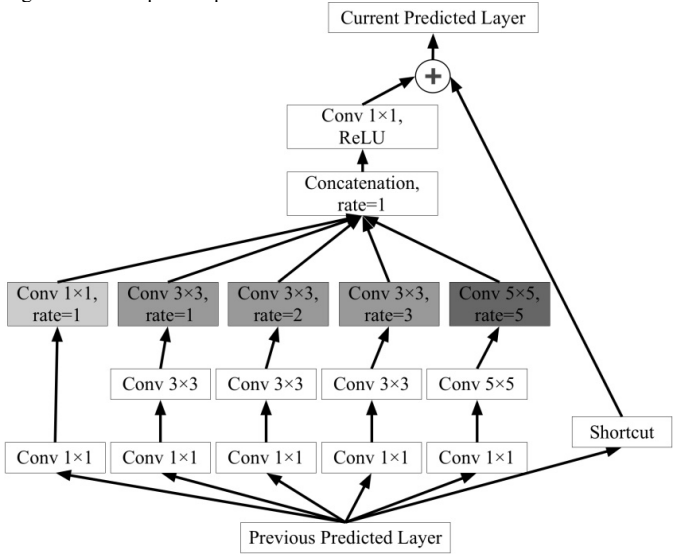

Fig. 3 The structure of multiple receptive field blocks (MRF)

For example, as shown in Fig. 2, the 1×1 convolution layer with dilated ratio 1 only focus on a center point's pixel of object, the 3×3 convolution layers with dilated ratios 1, 2 and 3 pay close attention to 9 pixels around one object at different spatial location. What's more, the 5×5 convolution layer with dilated ratio 5 has a larger receptive field, and more contextual information are introduced. According to these designed convolution layers, more positions of one object and its adjacent background are focused. In addition, the different convolution layers will learn different weights according to different receptive fields, and it has two advantages. On the one hand, the bigger weight will be learned according to smaller receptive fields and it will be assigned to these pixels whose positions are nearer to the center of object. It shows that these pixels are more important than the further ones. On the other hand, more contextual information is also focused by larger receptive field, and smaller weights are allocated to these further pixels. Thus, the MRF can capture the characteristic of different position pixels of one object and its adjacent background with different weights, which is helpful for extracting high quality features.

The detail structure of MRF is shown in Fig. 3, firstly, we adopt one bottleneck layer (1×1 convolution layer) to decrease and unify the number of channels in the previous feature map. Secondly, several convolution layers with different kernel sizes such as 1×1 convolution layer, 3×3 convolution layer, 5×5 convolution layer and several dilated convolution layers with different dilated ratios e.g. 3×3 convolution layer with dilated rate 1, 3×3 convolution layer with dilated rate 2, 3×3 convolution layer with dilated rate 3 are followed. Finally, the feature maps of all convolution layers with different receptive fields are concatenated to be inputted into a 1×1 convolution layer. In addition, we also design a shortcut layer likely ResNet [38] to maintain the performance of pervious layer.

### C. Small-Object-Focusing Weakly-supervised Segmentation Module (SWS)

Small-object-focusing weakly-supervised segmentation module (SWS) is a weakly supervised semantic segmentation at the level of bounding box. The input of SWS is the feature conv3_3_E. The feature map conv3_3_E shown in Fig. 1 is generated by FPN [16], which contains both semantic information and detailed information. As shown in Fig. 4, the ground truth of weakly-supervised segmentation is generated by only focusing the small object's bounding box. The output of SWS is the predicted score maps corresponding to foreground class and background class, respectively. The details of small-object-focusing weakly-supervised segmentation module are as follows.

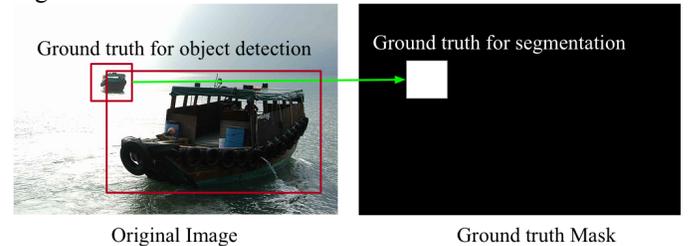

Fig. 4 The ground truth of small-object-focusing weakly-supervised semantic segmentation module

Small objects are only focused on the semantic segmentation branch without considering the class of object, and pixels of the ground truth mask are labeled as foreground if it falls within the bounding box of small objects. We define two thresholds $T_1$ and $T_2$ for limiting the areas of small objects. The objects whose areas are in $[T_1, T_2]$ are labeled as foreground. We also define some pixels as invalid pixels according to the setting threshold $T_1$, and the objects whose areas are lower than $T_1$ are overlooked in the training phase. The reason is that the extreme small object has not enough effective information for training segmentation module, which will result in incorrect prediction. The objects whose areas are higher than in $T_2$ are labeled as background. If a pixel is overlapped with multiple ground truths such as foreground truth and background truth and invalid truth, foreground truth will be selected preferentially. An example of ground truth for semantic segmentation is shown in Fig. 4.

The weakly-supervised semantic segmentation module consists of two transposed convolution layers following *conv3_3_E* and one transition convolutional layer with a ReLU non-linearity layer. Finally, a binary softmax classifier is designed to predict probability score of each pixel in the output mask as shown in Fig. 1.

*D. The Training of Multiple Tasks*

During the process of training, our final loss function includes the binary cross-entropy loss for semantic segmentation $L_{seg}(I, M)$, the softmax loss for classification $L_{conf}(I, C)$ and the smooth-l1 loss function for localization $L_{loc}(I, B)$. In short, the final loss combines detection loss with semantic loss, which is expressed as follows.

$$L(x, l, c, g, I, M) = L_{det}(x, l, c, g) + \alpha L_{seg}(I, M) \quad (1)$$

$$L_{det}(l, x, c, g) = \frac{1}{N}\left(L_{conf}(x, c) + \beta L_{loc}(x, l, g)\right) \quad (2)$$

where $L_{det}(x, l, c, g)$ is the detection loss, $L_{seg}(I, M)$ is the segmentation loss, $\alpha$ is a balance factor between detection task and segmentation task, $\beta$ is another balance factor between classification and localization in detection. *I* is the input image, and *M* is the ground truth at the level of bounding box for segmentation. *g* is the ground truth box for detection, *c* is the confidence score for predicted bounding box, *l* is the predicted bounding box, and *x* represents the predicted bounding box whether matches the ground truth or not.

Specifically, the main detection loss function is shown in (3) and (4), which is similar with the loss function of original SSD [19].

$$L_{conf}(x, c) = -\sum_{i \in Pos}^{N} x_{ij}^p log(\hat{c}_i^p) - \sum_{i \in Neg} log(\hat{c}_i^0) \quad (3)$$

$$L_{loc}(x, l, g) = -\sum_{i \in Pos}^{N} \sum_{m \in (cx, cy, w, h)} x_{ij}^k smooth_{L1}(l_i^m - \hat{g}_j^m) \quad (4)$$

Where $\hat{c}_i^p = \frac{exp(e_i^p)}{\sum_p exp(e_i^p)}$, $\hat{g}_j^{cx} = (g_j^{cx} - d_i^{cx})/d_i^w$, $\hat{g}_j^{cy} = (g_j^{cy} - d_i^{cy})/d_i^h$, $\hat{g}_j^w = log\left(\frac{g_j^w}{d_i^w}\right)$ and $\hat{g}_j^h = log\left(\frac{g_j^h}{d_i^w}\right)$. *d* is designed default bounding box. Similar to SSD, we regress offsets for the center $(cx, cy)$ of the default bounding box (*d*) and for its width (*w*) and height (*h*).

The added loss function of auxiliary segmentation task is shown in (5).

$$L_{seg}(I, M) = -\frac{1}{WH}\left(\sum_{w,h} log(Y_{w,h}, M_{w,h})\right) \quad (5)$$

Where $Y \in [0, 1]^{1 \times W \times H}$ represents the predicted mask, $M \in [0, 1]^{1 \times W \times H}$ is the bounding-box-level ground truth for segmentation. During the training phase, the two tasks are treated fairly. The balance factor α and β are set to 1.

## IV. EXPERIMENTS

The extensive experiment is conducted on two main detection datasets, which are PASCAL VOC [41] and MS COCO [42]. For PASCAL VOC, we use the common dataset division for evaluation. During the training phase, we use the VOC2007 trainval and VOC2012 trainval as training data. The VOC2007 test is used as testing data. The mean accuracy precision (mAP) with a fixed value of Intersection over Union (IoU) is used as evaluation function. A predicted bounding box is positive if the value of IoU with ground truth is higher than 0.5. For MS COCO, we follow a popular split, which uses train2015 as training dataset and minval2015 as testing data.

The baseline of detector choses the classic based-VGG16 SSD due to its higher accuracy and faster running speed, and we can confirm the effectiveness of the proposed modules on our detector by combining the classic based-VGG16 SSD on different datasets. Specifically, there are seven detection features including five MRFs on our detector MRFSWSnet300, and there are eight detection features including six MRFs on our detector MRFSWSnet512. The small-object-focusing weakly-supervised segmentation module (SWS) is integrated into detection branch following *conv3_3_E* layer for both MRFSWSnet300 and MRFSWSnet512. We follow the SSD training strategy throughout our experiments.

Our detector is also compared with some state-of-the-art methods including Faster RCNN [14], R-FCN [15], FPN [16], Yolov2 [25], SSD [19], DSSD [20], RFB [34], RetinaNet500 [23], these results of state-of-the-art detector are from original official published expect for RFB300 and RFB512. RFB300 and RFB 512 are trained and tested by ourselves on the same 4 Titan X GPUs for fair comparison.

*A. Experiment on PASCAL VOC*

On PASCAL VOC dataset, we do the training on a machine with 4 Titan X GPUs. The training process is little different

TABLE I
THE PERFORMANCE OF DIFFERENT DETECTORS ON PASCAL VOC2007 *TEST*

| Method | backbone | Time | mAP (%) |
|---|---|---|---|
| Faster RCNN [14] * | VGG16 | 147ms | 73.2 |
| Faster RCNN ++ [27] * | ResNet-101 | 3.36s | 76.4 |
| R-FCN [15] * | ResNet-101 | 110ms | 80.5 |
| YOLOv2 [25] * | darknet | 25ms | 78.6 |
| SSD300 [19] * | VGG16 | **12ms** | 77.2 |
| SSD512 [19] * | VGG16 | 28ms | 79.8 |
| DSSD513 [20] * | ResNet-101 | 182ms | **81.5** |
| RFB300 [34] | VGG16 | **15ms** | 79.9 |
| RFB512 [34] | VGG16 | 30ms | **81.5** |
| MRFSWSnet300 | VGG16 | **15ms** | 80.9 |
| MRFSWSnet512 | VGG16 | 31ms | **81.8** |

s represents second, and ms is millisecond. * represents that results are from original reference.

from original SSD [19]. At the beginning of training, the learning rate is a "warm up" process, whose range is from $10^{-6}$ to $10^{-4}$ at the first 10 epochs. After 10 epochs, the learning rate is set to $10^{-4}$ until 150-*th* epochs. Then, it is divided by 10 at 150-*th* epoch and 250-*th*. The total number of training epochs is set to 300. Following original SSD, the weight decay is 0.0005, the momentum is 0.9, and the batch size is set to 32. These parameters are similar with the original SSD. We use the pre-trained VGG-16 on the ILSVRC CLS-LOC dataset [40] to initialize our detector. $T_1$ and $T_2$ are set as 1024 and 9216. Its fc6 layer and fc7 layer in original VGG-16 are instead of convolution layers with the sub-sampling parameters, and the fc8 layer is removed. Other new convolutional layers of our detector including layers of auxiliary segmentation branch are

TABLE II
THE PERFORMANCE OF DIFFERENT DETECTORS ON MS COCO 2015

| Method | backbone | Time | Avg. Precision IOU: | | | Avg. Precision/Area | | |
|---|---|---|---|---|---|---|---|---|
| | | | 0.5:0.95 | 0.5 | 0.75 | S | M | L |
| Faster RCNN [14] | VGG16 | 147ms | 24.2 | 45.3 | 23.5 | 7.7 | 26.4 | 37.3 |
| Faster RCNN ++ [27] | ResNet-101 | 3.36s | 34.9 | 55.7 | 37.4 | 15.6 | 38.7 | **50.9** |
| FPN [16] | ResNet-101-FPN | 240ms | **36.2** | **59.1** | **39.0** | 18.2 | 39.0 | 48.2 |
| R-FCN [15] | ResNet-101 | 110ms | 29.9 | 51.9 | | 10.8 | 32.8 | 45.0 |
| Mask RCNN [26] | ResNext-101-FPN | 210ms | **37.1** | **60.0** | **39.4** | 16.9 | **39.9** | **53.5** |
| YOLOv2 [25] | darknet | 25ms | 21.6 | 44.0 | 19.2 | 5.0 | 22.4 | 35.5 |
| SSD300 [19] | VGG16 | **13ms** | 25.1 | 43.1 | 25.8 | -- | -- | -- |
| SSD512 [19] | VGG16 | 28ms | 28.8 | 48.5 | 30.3 | -- | -- | -- |
| DSSD513 [20] | ResNet-101 | 182ms | 33.2 | 53.3 | 35.2 | 13.0 | 35.4 | **51.1** |
| RetinaNet500 [23] | ResNet-101-FPN | 90ms | 34.4 | 53.1 | 36.8 | 14.7 | 38.5 | 49.1 |
| RetinaNet800 [23] | ResNet-101-FPN | 198ms | **39.1** | 59.1 | **42.3** | 21.8 | 42.7 | 50.2 |
| RFB300 | VGG16 | **16ms** | 30.3 | 49.3 | 31.8 | 11.8 | 31.9 | 45.9 |
| RFB512 | VGG16 | 33ms | 33.0 | 52.7 | 34.7 | 15.9 | 37.7 | 47.9 |
| MRFSWSnet300 | VGG16 | **16ms** | 30.5 | 49.8 | 31.7 | 14.3 | 32.3 | 45.3 |
| MRFSWSnet512 | VGG16 | 28ms | 33.1 | 53.0 | 34.7 | **17.5** | 37.1 | 47.9 |

s represents second, and ms is millisecond.

initialized with the MSRA method [43]. Table I shows the performance of different detectors on PASCAL VOC2007 test, and the bond font in Table I represents the best three results.

As shown in Table I, our detector MRFSWSnet outperforms original SSD on both resolution settings. The mAP of MRFSWSnet300 is 3.7% higher than original SSD300, meanwhile it also keeps almost equivalent running speed. With the high resolution, mAP of MRFSWSnet512 improves 2% from 79.8% to 81.8% while keeping the real-time speed same as SSD512. Compared with other state-of-the-art detectors such as popular state-of-the-art two-stage methods R-FCN [15] and detector combined with receptive fields block RFB [34], our detector still shows a significant performance improvement. It is also better than most one stage and two stage object detection systems equipped with very deep base backbone networks such as ResNet101.

*B. Experiment on MS COCO*

The strategy for training our detector on MS COCO is almost similar with the strategy on PASCAL VOC. However, the defaulting anchors are slight different from anchors on PASCAL VOC2007 to fit COCO dataset, whose scale is smaller. The update strategy of learning rate also adopts "warm up" at the first 10 epochs. After 10-*th* epochs, the learning rate is set to $2 \times 10^{-4}$ until 150-th epochs. Then, it is divided by 10 at 200-*th* epoch and 300-*th* epoch. The total epoch is set to 350. The momentum is set to be 0.9 and the weight decay is set to be 0.0005, which are consistent with the original SSD settings. $T_1$ and $T_2$ are set as 1024 and 9216. Similar to the experiment used for PASCAL VOC2007, we use the pre-trained VGG-16 on the ILSVRC CLS-LOC dataset [40] to initialize our detector. Table II shows the performance of different detectors on COCO 2015 dataset, the bond font in Table II represents the best three results.

From the Table II, the best three methods with higher mAP such as 36.2%, 37.1% and 39.1% are all region-based two-stage detector, but whose running time are much slower. Compared with these state-of-the-art methods, the average precision of MRFSWSnet300 can achieve 30.5%, and the mAP at IoU with 0.5 is 49.8% at lower resolution on the COCO 2015 dataset, but the running time of MRFSWSnet300 is much faster than these methods, and it is much better than the baseline detector SSD300. With lower resolution, MRFSWSnet300 is even better than that of R-FCN [15] with higher input size (600×1000), which employed ResNet-101 as the backbone under the two stages framework. The speed of our detector is only 16 milliseconds and 7 times faster than the speed of R-FCN.

In aspect of the bigger model, the result of MRFSWSnet512 with higher resolution (512×512) is also better than SSD512, the relative improvement of average precision is 14.9% from 28.8% to 33.1%. Our detector with higher resolution only consumes 28 milliseconds per frame, which is equivalent to SSD512. Compared with the recent advance state-of-the-art one-stage model RetinaNet500, the result of our detector is little inferior with the similar input size. However, RetinaNet is based on deeper and more complicated ResNet101 with FPN backbone, and a new focus loss focusing on hard examples. MRFSWSnet512 is only based on lightweight VGG model with FPN backbone. Moreover, RetinaNet is not a real-time detector, which runs 90 milliseconds per frame. The speed of RetinaNet is far slower than MRFSWSnet512 with similar input size. Considering the precision and the running time, our detector achieves the state-of-the-art real-time detector.

In addition, our detector is also compared with RFB, whose architecture is similar with ours. The average precision of our detector is higher than RBF regardless of the size of input and the mAP of our detector is also higher than RFB at higher IOU evaluation criteria (0.75). What's more, the average precision of our detector on small object is far better than RFB. For example, the average precision of our detector on small object is 14.3% and 17.5% compared with 11.8% and 15.9% of RFB with input size of 300×300 and 512×512, respectively. Compared with the state-of-the-art detector RetainNet500, the precision MRFSWSnet512 of for small object detection is higher 2.8% with the similar input. This proves the effectiveness of the proposed MRFSWSnet for small object detection.

## C. Ablation Experiments

TABLE III
THE PERFORMANCE OF OUR DETECTOR WITH VARIOUS DESIGNS ON THE PASCAL VOC2007 TEST

| Method | Original | MRFSWSnet300 | | | | |
|---|---|---|---|---|---|---|
| Added RFB [34] | × | √ | × | × | × | |
| Added MRF | × | × | √ | √ | √ | √ |
| More anchors with | × | × | × | √ | √ | √ |
| All-object-focusing weakly-supervised segmentation (AWS) | × | × | × | × | √ | × |
| Small-object-focusing weakly-supervised segmentation (SWS) | × | × | × | × | × | √ |
| mean Average Precision (mAP)/(%) | 77.2 | 79.6 | 80.1 | 80.3 | 80.7 | **80.9** |

To further prove the effectiveness of the proposed modules including MRF and SWS on our detector, some experimental results with different settings on PASCAL VOC2007 test dataset are shown in Table III. The resolution for all experiments is same with 300×300.

*1) Multiple receptive field block (MRF)*

In order to better understand the effectiveness of MRF, we compared the original SSD without MRF [19], SSD with RFB [34] and SSD with MRF. The result is summarized in Table III. As shown in Table III, the mean average precision (mAP) of original SSD [19] with the data augmentation achieves 77.2%. With RFB architecture [34], the mAP of SSD obtains 79.6%, which is 2.4% superior than original SSD. Instead of RFB, the mAP of SSD with MRF achieves 80.1%, which is 2.9% higher than original SSD and 0.5% higher than RFB. This shows that MRF is helpful for improving the detection accuracy.

*2) More detection feature with more anchors*

The original SSD consists of only six predicted features for object detection. Recent research [44] show that the higher resolution feature has benefit to detect small objects. We thus add a new convolution layer *conv3_3_E* as predicted feature. With the finest resolution layer, more anchors are proposed to cover more small instances. The experimental result is shown in Table III. We can see that the performance of our detector with conv3_3_E can improve 0.2% from 80.1% to 80.3% on PASCAL VOC2007 test.

*3) Small-object-focusing weakly-supervised semantic segmentation module (SWS)*

To further verify the effectiveness of the proposed small-object-focusing weakly-supervised semantic segmentation module, we conduct two experiments with different setting on PASCAL VOC2007 test. In the first experiment, we add all-object-focusing weakly-supervised semantic segmentation module (AWS) into our detector. In other words, all objects are need to be identify in the segmentation module regardless of the size of the object. The improvement of object detection's performance is 0.4% from 80.3% to 80.7%. We believe that the weakly-supervised semantic segmentation module is important to improve the performance of object detection. In the second experiment, small-object-focusing weakly-supervised semantic segmentation module (SWS) is integrated into our detector which only focuses on small object as described in Section III.C.

As shown in Table III, the performance of object detection improves 0.6% compared with our detector without SWS, which is also 0.2% better than our detector with AWS. This indicates that the small-object-focusing weakly supervised semantic segmentation module for auxiliary training is crucial to enhance the performance of detection.

## D. Inference Speed Comparisons

TABLE IV
THE MAP AND SPEED OF VARIOUS DETECTORS ON THE MS COCO DATASET

| Method | backbone | Time(ms) | mAP(%) |
|---|---|---|---|
| Faster RCNN [14] | VGG16 | 147 | 24.2 |
| FPN [16] | ResNet-101-FPN | 240 | **36.2** |
| R-FCN [15] | ResNet-101 | 110 | 29.9 |
| Mask RCNN [26] | ResNext-101-FPN | 210 | **37.1** |
| YOLOv2 [25] | darknet | 25 | 21.6 |
| SSD300 [19] | VGG16 | **13** | 25.1 |
| SSD512 [19] | VGG16 | 28 | 28.8 |
| DSSD513 [20] | ResNet-101 | 182 | 33.2 |
| RetinaNet500 [23] | ResNet-101-FPN | 90 | 34.4 |
| RetinaNet800 [23] | ResNet-101-FPN | 198 | **39.1** |
| RFB300 | VGG16 | **16** | 30.3 |
| RFB512 | VGG16 | 33 | 33.0 |
| MRFSWSnet300 | VGG16 | **16** | 30.5 |
| MRFSWSnet512 | VGG16 | 28 | 33.1 |

ms is millisecond. * represents that results are from original reference.

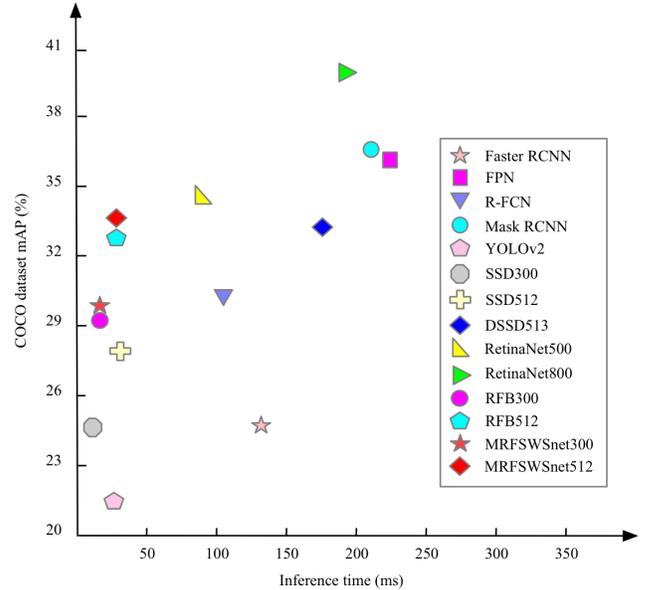

Fig. 5. The mAP vs speed of different detectors on MS COCO dataset

The mAP and running time comparison between our proposed MRFSWSnet and recent state-of-the-art detectors are shown in Fig. 5 and Table IV. We follow [23] to plot the speed/accuracy trade-off curve for some recent methods and our detector on the MS COCO dataset. This curve shows the relation between precision and inference time for each detector. The bond font in Table IV represents the best three results. From the Table IV, the best three methods with higher mAP such as 36.2%, 37.1% and 39.1% are all region-based two-stage detector, whose running time are much slower.

Our detector MRFSWSnet300 is little slower than original VGG16-based SSD300, but mAP of MRFSWSnet300 is much higher than SSD300. Meanwhile, it is faster than other real-time detectors such as YOLOv2, RFB300 with the same input size. With higher resolution (512×512), our detector MRFSWSnet512 also keeps lower running time (28ms per frame) and higher precision (33.1%). Considering the precision and running time, our detector is the state-of-the-art method among real-time detectors.

## V. CONCLUSION

In this paper, we propose a multiple receptive field and small-object-focusing weakly-supervised segmentation network (MRFSWSnet) for fast object detection. In MRFSWSnet, multiple receptive fields (MRF) are used to pay attention to different spatial locations of the object and its adjacent background with different weights. MRF can effectively enhance the feature discriminability and robustness according to the verification of extensive experiments. In addition, in order to further improve the accuracy of small object detection, a small-object-focusing weakly-supervised segmentation module (SWS) is integrated into the detection network for auxiliary training. SWS shows significant performance boosting for small object detection, while it still keeps real-time speed. Extensive experiments show the effectiveness of our detector on PASCAL VOC and MS COCO detection datasets.

REFERENCES

[1]  S. Minaeian, J. Liu et al., "Vision-based target detection and localization via a team of cooperative UAV and UGVs," *IEEE* Trans. Syst. *Man*. Cy-S., vol. 46, no. 7, pp. 1005-1016, Jul. 2016.
[2]  X. Yuan, X. Cao, X. Hao et al., "Vehicle detection by a context-aware multichannel feature pyramid," *IEEE* Trans. Syst. *Man*. Cy-S., vol. 47, no. 7, pp. 1348-1357, Jul. 2017.
[3]  A. B. Mabrouk, E. Zagrouba, "Abnormal behavior recognition for intelligent video surveillance systems: A review," Expert Syst. Appl., vol. 91, pp. 480-491, Jan. 2018.
[4]  S. Bashbaghi, E. Granger, R. Sabourin et al., "Deep learning architectures for face recognition in video surveillance," arXiv preprint arXiv:1802.0999, Jun. 2018, pp. 1-9.
[5]  W. Yang, B. Fang, Y. Tang., "Fast and accurate vanishing point detection and its application in inverse perspective mapping of structured road," *IEEE* Trans. Syst. *Man*. Cy-S., vol. 48, no. 5, pp. 755-766, May. 2018.
[6]  H. Kuang, L. Chen, L. Chan et al., "Feature selection based on tensor decomposition and object proposal for night-time multiclass vehicle detection," *IEEE* Trans. Syst. *Man*. Cy-S., vol. 49, no. 1, pp. 71-80, Jan. 2019.
[7]  N. Dalal, B. Triggs, "Histograms of oriented gradients for human detection," in Proc. IEEE Conf. Comput. Vis. Pattern Recognit., CA, USA, Jun. 2005, pp. 886-893.
[8]  D. G. Lowe, "Object recognition from local scale-invariant features," in Proc. IEEE Conf. Comput. Vis., Corfu, Greece, Sep. 1999, pp. 1150-1157.
[9]  K. E. A. Van de Sande, J. R. R. Uijlings, T. Gevers et al., "Segmentation as selective search for object recognition," in Proc. IEEE Conf. Comput. Vis., Barcelona, Spain, Nov. 2011, pp. 1879-1886.
[10] J. Pont-Tuset, P. Arbelaez, J. T. Barron et al., "Multiscale combinatorial grouping for image segmentation and object proposal generation," IEEE Trans. Pattern Anal. Mach. Intell., vol. 39, no. 1, pp. 128-140, Jan. 2017.
[11] C. L. Zitnick, P. Dollár, "Edge boxes: Locating object proposals from edges," in Proc. Eur. Conf. Comput. Vis., Zurich, Switzerland, Sep. 2014, pp. 391-405.
[12] K. He, X. Zhang, S. Ren, J. Sun, "Spatial pyramid pooling in deep convolutional networks for visual recognition," IEEE Trans. Pattern Anal. Mach. Intell., vol. 37, pp. 1904-1916, Jan. 2015.
[13] R. Girshick, "Fast R-CNN," in Proc. IEEE Conf. Comput. Vis., Santiago, Chile, Oct, 2015, pp. 1440-1448.
[14] S. Ren, K. He, R. Girshick. "Faster R-CNN: Towards real-time object detection with region proposal networks," in Proc. Int. Conf. Neural Inf. Process. Sys., Palais des Congrès de Montréal, Montréal CANADA, Sep. 2015, pp. 91-99.
[15] J. Dai, Y. Li, K. He et al., "R-FCN: Object detection via region-based fully convolutional networks," in Proc. Int. Conf. Neural Inf. Process. Sys., Barcelona, Spain, Dec. 2016, pp. 379-387.
[16] T. Y. Lin, P. Dollár, R. Girshick et al., "Feature pyramid networks for object detection," IEEE Conf. Comput. Vis. Pattern Recognit., Hawaii, USA, Jul. 2017, pp. 2117-2125.
[17] B. Singh, L. S. Davis, "An analysis of scale invariance in object detection–SNIP," in Proc. IEEE Conf. Comput. Vis. Pattern Recognit., Salt Lake City, Utah, USA, Jun. 2018, pp. 3578-3587.
[18] J. Redmon, S. Divvala, R. Girshick, "You only look once: Unified, real-time object detection." in Proc. IEEE Conf. Comput. Vis. Pattern Recognit., Las Vegas, Nevada, Jun. 2016, pp. 779-788.
[19] W. Liu, D. Anguelov, D. Erhan. "SSD: Single shot multibox detector," in Proc. Eur. Conf. Comput. Vis., Amsterdam, Netherlands, Sep. 2016, pp. 21-37.
[20] C. Y. Fu, W. Liu, A. Ranga et al., "DSSD: Deconvolutional single shot detector," in Proc. IEEE Conf. Comput. Vis. Pattern Recognit., Hawaii, USA, Jul. 2017, pp. 1-9.
[21] J. Ren, X. H. Chen, J. B. Liu et al., "Accurate single stage detector using recurrent rolling convolution," in Proc. IEEE Conf. Comput. Vis. Pattern Recognit., Hawaii, USA, Jul. 2017, pp. 1-9.
[22] Z. Shen, Z. Liu, J. Li et al., "DSOD: Learning deeply supervised object detectors from scratch," in Proc. IEEE Conf. Comput. Vis., Venice, Italy, Oct. 2017, pp. 7.
[23] T. Y. Lin, P. Dollár, R. Girshick et al., "Focal loss for dense object detection," in Proc. IEEE Conf. Comput. Vis., Venice Italy, Oct. 2017, pp. 1-10.
[24] S. Zhang, L. Wen, X. Bian et al., "Single-shot refinement neural network for object detection," in Proc. IEEE Conf. Comput. Vis. Pattern Recognit., Salt Lake City, Utah, USA, Jun. 2018, pp. 4203-4212.
[25] J. Redmon, A. Farhadi. "YOLO9000: better, faster, stronger," in Proc. IEEE Conf. Comput. Vis. Pattern Recognit., Hawaii, USA, Jul. 2017, pp. 7263-7271.
[26] C. Zhu, R. Tao, K. Luu et al., "Seeing small faces from robust anchor's perspective," in Proc. IEEE Conf. Comput. Vis. Pattern Recognit., Utah, USA, Jun. 2018, pp. 5127-5136.
[27] H. Levi, S. Ullman. "Efficient coarse-to-fine non-local module for the detection of small objects, arXiv preprint arXiv:1811.12152, 2018.
[28] C. Szegedy, W. Liu, Y. Jia et al., "Going deeper with convolutions," in Proc. IEEE Conf. Comput. Vis. Pattern Recognit., Boston, USA, Jun. 2015, pp. 1-9.
[29] C. Szegedy, V. Vanhoucke, S. Ioffe et al., "Rethinking the inception architecture for computer vision," IEEE Conf. Comput. Vis. Pattern Recognit., Las Vegas, Nevada, Jun. 2016, pp. 2818-2826.
[30] C. Szegedy, S. Ioffe, V. Vanhoucke et al., "Inception-v4, inception-resnet and the impact of residual connections on learning," in Proc. Thirty-First AAAI Conf. Artif. Intell., San Francisco, California USA, Feb. 2017, pp. 4-12.
[31] L. C. Chen, G. Papandreou, I. Kokkinos et al., "Deeplab: Semantic image segmentation with deep convolutional nets, atrous convolution, and fully connected crfs," IEEE Trans. Pattern Anal. Mach. Intell., vol. 40, no. 4, pp. 834-848, Apr. 2018.
[32] J. Dai, H. Qi, Y. Xiong et al., "Deformable convolutional networks," in Proc. IEEE Conf. Comput. Vis., Venice Italy, Oct. 2017, pp. 764-773.
[33] Z. Li, C. Peng, G. Yu et al., "Detnet: Design backbone for object detection," in Proc. Eur. Conf. Comput. Vis., Munich, Germany, Sep. 2018, pp. 339-354.
[34] S. Liu, D. Huang. "Receptive field block net for accurate and fast object detection," in Proc. Eur. Conf. Comput. Vis., Munich, Germany, Sep. 2018, pp. 385-400.
[35] S. Gidaris, N. Komodakis. "Object detection via a multi- region and semantic segmentation-aware cnn model," in Proc. IEEE Conf. Comput. Vis., Santiago, Chile, Oct, 2015, pp. 1134-1142.
[36] Z. Zhang, S. Qiao, C. Xie et al., "Single-shot object detection with enriched semantics," in Proc. IEEE Conf. Comput. Vis. Pattern Recognit., Salt Lake City, Utah, USA, Jun. 2018, pp. 1-9.
[37] K. He, G. Gkioxari, P. Dollár et al., "Mask r-cnn," in Proc. IEEE Conf. Comput. Vis., Venice Italy, Oct. 2017, pp. 2980-2988.
[38] K. He, X. Zhang, S. Ren et al., "Deep residual learning for image recognition," in Proc. IEEE Conf. Comput. Vis. Pattern Recognit., Las Vegas, Nevada, Jun. 2016, pp. 770-778.

[39] K. Simonyan, A. Zisserman. "Very deep convolutional networks for large-scale image recognition," in Proc. Int. Conf. Learn. Rep., San Diego, CA, May, 2015, pp. 1-14.
[40] O. Russakovsky, J. Deng, H. Su et al., "Imagenet large scale visual recognition challenge," Int. J. Comput. Vis., vol. 115, no. 3 pp. 211-252, Apr. 2015.
[41] M. Everingham, L. VanGool et al., "The pascal visual object classes (voc) challenge," Int. J. Comput. Vis., vol. 88, no. 2, pp. 303-338, Sep. 2009.
[42] T. Y. Lin, M. Maire, S. Belongie et al., "Microsoft coco: Common objects in context," in Proc. Eur. Conf. Comput. Vis., Switzerland, Zurich, Sep. 2014, pp. 740-755.
[43] K. He, X. Zhang, S. Ren et al., "Delving deep into rectifiers: Surpassing human-level performance on imagenet classification," in Proc. IEEE Conf. Comput. Vis., Santiago, Chile, Oct, 2015, pp. 1026-1034.
[44] P. Hu, D. Ramanan, "Finding tiny faces," in Proc. IEEE Conf. Comput. Vis. Pattern Recognit., Hawaii, USA, Jul. 2017, pp. 1522-1530.